# ADEQA: A Question-Answer based approach for joint ADE-Suspect Extraction using Sequence-To-Sequence Transformers

**Vinayak Arannil**
AWS AI

**Tomal Deb**
AWS AI

**Atanu Roy**
AWS AI

## Abstract

Early identification of Adverse Drug Events (ADE) is critical for taking prompt actions while introducing new drugs into the market. These ADEs information are available through various unstructured data sources like clinical study reports, patient health records, social media posts, etc. Extracting ADEs and the related suspect drugs using machine learning is a challenging task due to the complex linguistic relations between drug – ADE pairs in textual data and unavailability of large corpus of labelled datasets. This paper introduces ADEQA, a question- answer(QA) based approach using quasi supervised labelled data and sequence-to-sequence transformers to extract ADEs, drug suspects and the relationships between them. Unlike traditional QA models, natural language generation (NLG) based models don't require extensive token level labelling and thereby reduces the adoption barrier significantly. On a public ADE corpus, we were able to achieve state-of-the-art results with an F1 score of 94% on establishing the relationships between ADEs and the respective suspects.

## 1 Introduction

Everyday hundreds of drugs are being introduced to the market. However, every drug has contraindications. A study conducted by (Hazell and Shakir, 2006) showed that 7000 deaths are being caused by Adverse Drug Events (ADE) annually. Organizations like the World Health Organization (WHO), the Food and Drug Administration (FDA), the European Medicines Agency (EMEA), and the Medicines and Healthcare products Regulatory Agency (MHRA) maintain a reporting system that enables individuals to spontaneously report the experienced adverse effects related to the use of medicines or healthcare products (Hazell and Shakir, 2006). Although these systems store the adverse event information in a structured format, a vast amount of information still remains in the unstructured textual data like clinical trial reports, patient health records, medical transcripts, social media posts, etc. It's a tedious process to have humans go through each of these documents and record the mentioned adverse events and the related suspect drugs.

With the advancements in machine learning, specifically in the field of Natural Language Processing (NLP), information extraction models are being widely used to extract useful information from unconstrained texts. Such models can learn contextual patterns to identify and extract specific entities, after being trained using large corpus of annotated data. Similar approaches have been applied to extract ADEs and suspect drugs using Named Entity Recognition models (Wikipedia contributors, 2023). However, since the ADEs are semantically similar to any other unrelated symptoms, most often such models predict false positives. Hence, improving precision in this task depends on the ability to contextually relate the ADEs to the relevant suspect drug(s), instead of extracting the ADEs independently. Unfortunately, generalized extraction of relationships among entities in <subject, predicate, object>form is still a challenging task in the NLP ecosystem. In this work, we wanted to address this shortcoming by modelling the Drug-ADE relationship extraction task as Question Answering tasks.

Deep Neural Network based, supervised NLP models require tens of thousands of annotated data to learn the contextual information and identify hidden patterns. For tasks like NER (Wikipedia contributors, 2023), annotation of text data is a critical prerequisite needing manual effort, coupled with domain knowledge. The classical approach of annotating entities with B-I-O offsets (Huang et al., 2015) increases the efforts further. Considering the ongoing exponential growth of data, annotating huge corpus of new data to train or retrain the models in future would be very expensive, if not entirely impossible. This is where models which require light weight labelling come into the rescue. Sequence to sequence transformer (Vaswani et al., 2017) models like T5 (Raffel et al., 2020) can be trained by transforming the entities and relationships to be extracted into text sequences, so that these models can learn the contextual patterns to identify and generate the required entities, without any explicit token offset information.

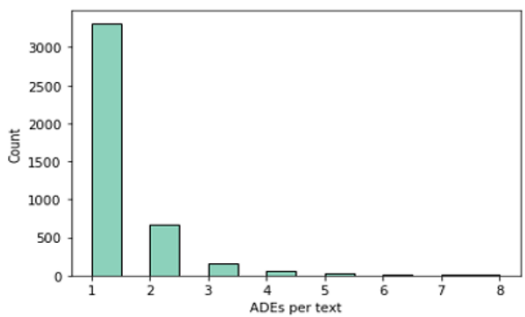 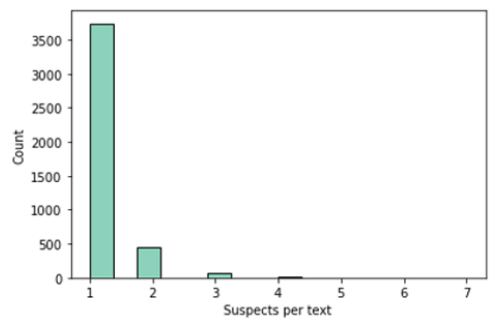

Figure 1: Distribution of counts of ADEs/text in the dataset

Figure 2: Distribution of counts of suspects/text in the dataset

In this paper, we intend to elaborate two approaches of modelling Drug – ADE relation extraction as Question-Answering solution using natural language generation (NLG) technique via sequence-to-sequence modelling. First approach is a two-step solution that first extracts Drugs and ADEs and subsequently confirms associations between them, while the second one directly discovers the potential Drug – ADE pairs from a given text. One of our approach achieved state-of-the-art F1 scores of 94% in establishing the relationships between ADEs and the respective suspects on the public ADE benchmark corpus (Gurulingappa H, 2012).

## 2 Related Works

Several works have already been tried out for ADE-suspect identification task. Earlier approaches focused mainly on pipeline design with a NER model to extract entities and their offsets, followed by a relation classification model which takes two pair of entities and identify the relation between them (Gurulingappa H, 2012)(Li and Ji, 2014). With the advancements in deep learning, RNN based sequence models like LSTMs (Hochreiter and Schmidhuber, 1997), GRUs (Cho et al., 2014) started being applied on all NLP use cases. (Li et al., 2016) used a feed forward neural network to jointly extract drug-disease entity mentions and their relations.

(Li et al., 2017) explored bidirectional LSTMs for learning entity representations from text sequences. They used Shortest Dependency Paths (SDP) between probable entities to identify related ADEs and suspects.

(Ramamoorthy and Murugan, 2018) proposed a self-attention-based Bi-LSTM model for facilitating intra-sequence interaction in the given text sequence. The same work conceptually considered ADE extraction as a question answering problem, where the text sequence becomes the context and the drug whose adverse effects are to be predicted, becomes the query. However, rather than selecting an answer (adverse effect) from a vocabulary, they consider each token in the sequence as a potential ADE and embed this logic directly into the modeling than really having QA model. This adds additional computational complexity. Several other studies were also conducted using bidirectional LSTMs for the ADE-suspect extraction task (Sorokin and Gurevych, 2017)(Henry et al., 2019)(Christopoulou et al., 2019)(Lample et al., 2016)(Yang et al., 2018).

Attention based models like transformers (Vaswani et al., 2017) which can learn contextual patterns efficiently have more or less replaced LSTM based models lately. (Wei et al., 2020)(Alimova and Tutubalina, 2020) applied pretrained BERT (Devlin et al., 2019) models for the ADE extraction task. (Wang and Lu, 2020) created shared layers between NER and RE model for joint ADE-suspect identification. Current state-of-the-art model on this task by Haq et al., (Haq et al., 2021) uses BioBERT (Lee et al., 2019) as the base in a NER-RE pipeline design with RE models placed sequentially after the NER model, and are fed the results of the NER model, the context, embeddings, and dependency tree for feature gen-eration to classify the relationships. Multi-turn QA (Li et al., 2019) also casts the NER-RE problem as a multi-turn question answering task. MRC4ERE (Zhao et al., 2020b) improves on this question answering approach by leveraging a diverse set of questions. However, both the approaches consider deterministic methods for extracting the answers and uses BERT (Devlin et al., 2019) for modeling.

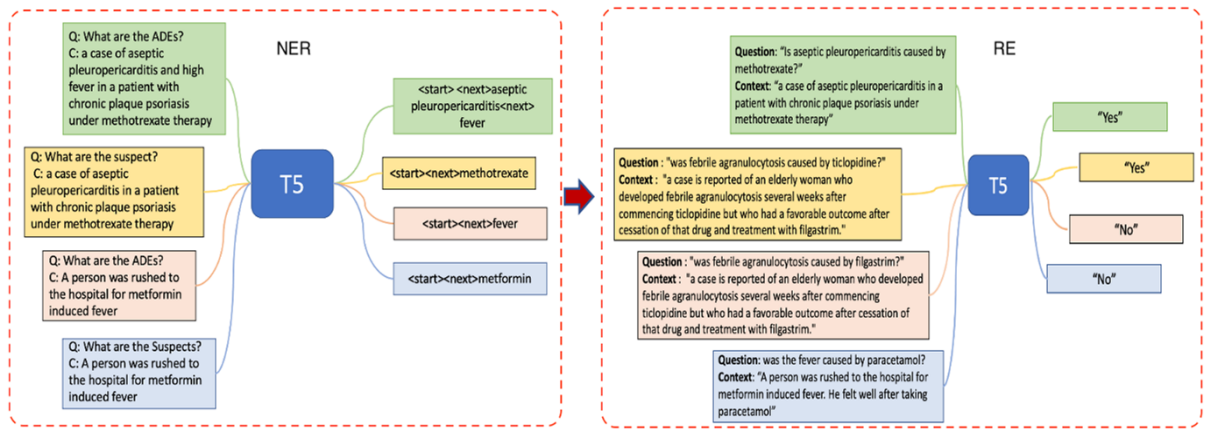

Figure 3: Approach 1: ADE, suspects and relations are identified separately using same model via multi-tasking.

Sequence-to-sequence transformer models like BART (Lewis et al., 2020), T5 (Raffel et al., 2020), etc. are being studied for ADE-suspect extraction task recently and shown positive results. Our work is heavily inspired from some of the latest researches which adopted NLG models for the relation extraction tasks. REBEL (Huguet Cabot and Navigli, 2021) which achieved state of the art results in multiple RE benchmark datasets transformed the entity relationships as text sequence of triplets and used BART (Lewis et al., 2020) to generate these triplets. Similarly, TANL by Paolini et al., 2021 (Paolini et al., 2021) frame this as a translation task by generating an augmented text with entity and relation information marked. (Raval et al., 2021) explored T5 model for medical product safety monitoring in social media.

## 3 Dataset

We have used the ADE dataset (Gurulingappa H, 2012), which is an annotated data of adverse events and drugs identified from biomedical texts. The original corpus is distributed in three files i.e., drug-ade relation data, drug-dosage relation data and ade-negative data, out of which we used the drug-ade data for our experiments.

Although there are begin and end offsets annotated for the ADEs and suspects, our approaches

**Algorithm 1** NER-RE pipeline using multitask learning

1: **Input:**
2: Q = question
3: C = context
4: **Output:**
5: A = answer
6: **Start:**
7: ADEs = []
8: Suspects = []
9: ade_sus = []
10: **for** every text **do**
11:     Q = "what are the ADEs?"
12:     C = text
13:     A = get_ades(Q, C)
14:     *example A=<Start>ade1<next>ade2<next>ade3*
15:     ADEs += [ade1,ade2,ade3]
16:     Q = "what are the suspects?"
17:     C = text
18:     A = get_suspects(Q, C)
19:     *example A=<Start>suspect1<next>suspect2*
20:     Suspects += [suspect1,suspect2]
21:     **for** ade in ADEs **do**
22:         **for** suspect in Suspects **do**
23:             Q = "is ade casused by suspect?"
24:             C = text
25:             A = confirm_association(Q, C)
26:             *example A = 'Yes' or 'No'*
27:             **if** A == 'Yes' **then**
28:                 ade_sus += [(ade,suspect)]
29:             **end if**
30:         **end for**
31:     **end for**
32: **end for**

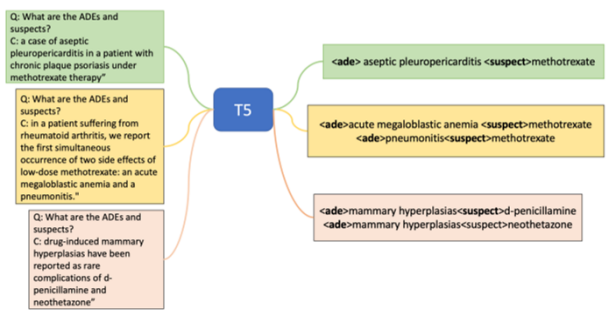

Figure 4: Approach 2: Joint end-to-end extraction as a single task

do not require that information. There are 6,821 texts available in the corpus with only 20% of them including more than one ADE or suspect, as seen in Fig. 1 and 2. Altogether, there are 2984 unique ADEs and 1050 unique suspects in the whole corpus. A sample row from the dataset looks like this "10030778|Intravenous azithromycin-induced ototoxicity.|ototoxicity|43|54|azithromycin|22|34". Columns 2, 3, and 6 provide the text, ADE, and suspect information, respectively.

## 4  Approaches

Sequence-to-sequence transformer models like T5 are capable of handling several NLP tasks concurrently. As explained in (Raffel et al., 2020), every NLP task we consider including translation, question answering, classification, etc. is cast as feeding the model text as input and training it to generate some target text (Raffel et al., 2020). This allows us to use the same model, loss function, hyperparameters, etc. across diverse set of tasks. In order to train a single model on the diverse set of tasks described above, T5 cast all of the tasks we consider into a "text-to-text" format that is, a task where the model is fed some text for context or conditioning and is then asked to produce some output text (Raffel et al., 2020). T5 framework provides a consistent training objective both for pre-training and fine-tuning. Specifically, the model is trained with a maximum likelihood objective (using "teacher forcing" (Williams and Zipser, 1989) regardless of the task (Raffel et al., 2020). To specify which task the model should perform, we add a task-specific (text) prefix to the original input sequence before feeding it to the model (Raffel et al., 2020). We use questions as the prefix in our experiments.

In this section we introduce two approaches for solving ADE- suspect extraction problem using

---

**Algorithm 2** Joint ADE-Suspect relation extraction as single task

1: **Input:**
2: Q = question
3: C = context
4: **Output:**
5: A = answer
6: **Start:**
7: ade_sus = [ ]
8: **for** every text **do**
9:    Q = "what are the ADEs and suspects?"
10:   C = text
11:   A = get_relations(Q, C)
12:   example A=<Start>
        ade1<next>suspect1<next>ade2<next>suspect2
13:   ade_sus+=[(ade1,suspect1),(ade2,suspect2)]
14: **end for**

---

sequence to sequence modeling.

### 4.1  NER-RE pipeline using multitask learning

A Named Entity recognition (NER) model followed by a Relation Extraction (RE) model is the conventional method to solve this task. In this approach, we first extract ADEs and suspects from the text independently. Then, link the ADEs and suspects using one-to-one mapping, to identify whether they are related or not. Although transformer (Vaswani et al., 2017) based models have been shown to learn patterns from the textual contexts, there is no guarantee that the extracted event or suspect is the actual adverse event or suspect. A Relation extraction module followed by the NER, can eliminate such false positives efficiently. In addition to improved performance, generative models like T5 (Raffel et al., 2020) require less annotation effort compared to traditional NER models which require data to be annotated in B-I-O format, which is time-consuming and inconvenient. For example, a sentence like "A man was rushed to the hospital for metformin induced severe fever" should be annotated as " O O O O O O O O B-SUS O B-ADE I-ADE". In real-life, most often we won't have such extensive annotated data which is necessary in order to use any standard NER and RE models (Haq et al., 2021). The algorithm for this approach is shown in Algorithm 1. Detailed explanation of the approach is available in section 5.

## 4.2 Joint ADE-Suspect relation extraction as single task

If we can transform the pair of ADEs and suspects and their relationship into a text sequence, we can use sequence to sequence models like T5 (Raffel et al., 2020), BART (Lewis et al., 2020), etc. to perform any language generation task end-to-end. In this second proposed approach, we use a single T5 model to perform the end-to-end extraction task. Specifically, extract ADEs, suspects, and their relationships all at once. This strategy fully exploits the learning capabilities of the T5 model. The algorithm to perform this task is shown in Algorithm 2. Section 5 provides an in-depth explanation of the strategy.

## 5 Experiments and Setup

For the first approach of NER-RE pipeline, as shown in Fig. 3, we used a single model to execute multi-task learning in order to identify suspects, ADEs, and examine the relationships between the two. This is equivalent to mapping an input sequence of n words to an output sequence of m ADEs or supects, conditioned over a question and a context as shown in (1) and (2). We employed questions such as "What are the ADEs?" and "What are the suspects?" for ADEs and suspects extraction, respectively. Since there can be multiple ADEs and suspects within the same text, the model should be able to generate all available entities. We used a special token <next>in the ground truth to teach the model to generate the next entity one after the other. Since <next>token can appear multiple times in the output, we removed the repetition penalty in the T5 model. Additional post processing was used to eliminate duplicate results.

$$p(y_{1i}^{ADE}, y_{2i}^{ADE} ..., y_{mi}^{ADE} \mid x_{seqi}^{Q}, x_{seqi}^{C}) \quad (1)$$

$$p(y_{1i}^{Suspect}, y_{2i}^{Suspect} ..., y_{mi}^{Suspect} \mid x_{seqi}^{Q}, x_{seqi}^{C}) \quad (2)$$

Relationship extraction module in the approach 1 was also modeled as a QA task. Initially, we trained the model by using questions like "what caused the <ADE>?" by providing the whole text as the context and allowing the model to predict the suspects directly. Although the model was performing well in identifying the suspects, it made mistakes when there are more than one drug names present in the text. Since it is difficult to derive an accurate confidence score from a seq-to-seq model like T5, we had to formulate some strategy to identify negative relations. To combat this, we created questions with binary responses that the model may produce. For example, given a context like "A person was rushed to the hospital due to metformin induced fever. He was feeling better after taking tylenol.", the questions were framed like "Was the fever caused by metformin?" and "Was the fever caused by tylenol?". In this way, the model was able learn and understand the context and provide 'Yes' or 'No' answers as shown in (3). We could discard the negative relationships using the 'No' output hence improving the overall precision. This would not have been possible if we had used traditional QA models. They provide deterministic results by outputting a phrase from the input text itself. However, NLG based models can generate answers which are not present in the input text.

$$p(y_i^{Yes|No} \mid x_{seqi}^{Q}, x_{seqi}^{C}) \quad (3)$$

In order to execute our second approach of end-to-end extraction, we used a single question like "what are the ADEs and suspects?" and tuple generation method similar to (Huguet Cabot and Navigli, 2021). We used additional tokens like <ade>, <suspect>to demarcate the tuples as shown in Fig. 4. It was also observed that, unlike (Huguet Cabot and Navigli, 2021) we didn't have to perform any entity sorting, based on the positions in the text. Model was able to perform the extractions of tuples accurately without sorting. This is equivalent to mapping an input sequence of n words to an output sequence of m pairs of ADEs and supects, conditioned over a question and a context as shown in (4).

$$p(y_{1i}^{(ADE-Suspect)}, ..., y_{mi}^{(ADE-Suspect)} \mid x_{seqi}^{Q}, x_{seqi}^{C}) \quad (4)$$

Models were trained using NVIDIA T4 GPUs with a batch size of 4. We used G4dn.xlarge instances of AWS which provides T4 GPUs. Additional hyperparameter tuning was also performed using baysian optimization (Nguyen, 2019). We have used input sequence length of 128 and target sequence length of 32. It took around 10-15 minutes to finetune a T5 model for a training set of 5500 texts with the remaining texts from the corpus used as the evaluation data. While evaluating the extracted ADEs and suspects, we considered

Table 1: Results comparison for both the approaches

| Entity | Approach1 F1* | Approach2 F1* |
|---|---|---|
| ADE | 0.91 | 0.89 |
| Suspect | 0.98 | 0.96 |
| Relationship | 0.95 | 0.83 |

*partial match micro

partial match along with strict match, as prediction or ground truth sometimes contain adjectives which goes missing in either side. For example, a text like "a man was rushed to the hospital for metformin induced severe fever". Here the ground truth might be just "fever", while the model would learn to predict "severe fever" or vice-versa. For the partial match calculation, we used levenshtein distance based distance computation between generated sequence and ground truth sequence.

## 6 Results

Comparative performance of both the approaches are shown in Table 1. Here we considered partial match F1 score to compare both the results. On strict evaluation, approach 1 found to be better than approach 2. Also, Table 2 shows the comparison of performance with the existing baselines. Approach 1 achieved state-of-the-art results on establishing the relationship between ADEs and suspects. Since RE is treated as a separate task in approach 1, we evaluated the RE performance independently without tying with the NER output. i.e. we checked if a pair of an event and a drug in the evaluation text are related by adverse effect or not. This gives the intuition that, even if the NER task predicts false results, we can effectively eliminate them using the subsequent RE model, hence improving the end-to-end precision. Fig. 5 shows the confusion matrix of the relation extraction task from approach 1.

Individual metrics for ADE, suspect and RE from the approach 2 are evaluated by splitting the result into ADEs and suspects and by checking if a pair of ADE-Suspects are related correctly or not. it's also observed that the approach 2 suffers when there are more than 3 pairs of ADEs and suspects with in a text. As seen in Fig. 6, identification of ADEs and suspects is most effective when there are 3–4 or less of these entities per text. Ratio of correct to wrong prediction goes up as the num-ber of entities per text increases. However, given that the majority of literature only mentions one to two ADEs or suspects in a text, as evidenced by the original data distribution (Fig. 1 and 2), this performance is ideal for real-life situations.

Table 2: Comparison with existing baselines (F1 score)

| Strict match | NER | RE |
|---|---|---|
| SpBERT (Eberts and Ulges, 2019) | 0.892 | 0.792 |
| CLDR+CLNER (Theodoropoulos et al., 2021) | 0.883 | 0.792 |
| Table sequence (Wang and Lu, 2020) | 0.897 | 0.801 |
| CMAN (Zhao et al., 2020a) | 0.894 | 0.837 |
| TANL () | 0.902 | 0.806 |
| TANL(multi-dataset) (Paolini et al., 2021) | 0.90 | 0.80 |
| TANL(multi-task) (Paolini et al., 2021) | 0.912 | 0.838 |
| REBEL (Huguet Cabot and Navigli, 2021) | - | 0.822 |
| Deeper (Crone, 2020) | 0.894 | 0.837 |
| SparkNLP (Haq et al., 2021) | 0.917 | 0.900 |
| **ADEQA (Approach 1)** | 0.885 | **0.945** |
| ADEQA (Approach 2) | 0.867 | 0.772 |
| Partial match | | |
| ADEQA (Approach 1) | 0.941 | 0.945 |
| ADEQA (Approach 2) | 0.924 | 0.829 |

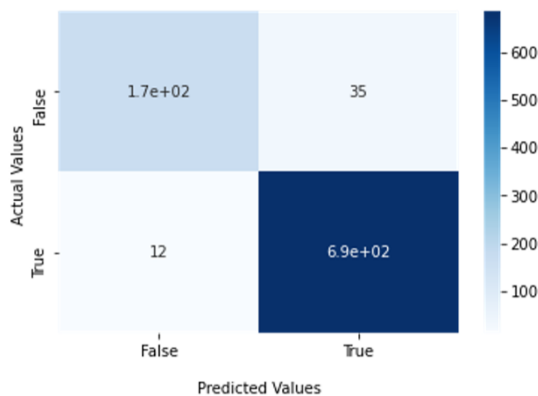

Figure 5: Confusion matrix of relation extraction

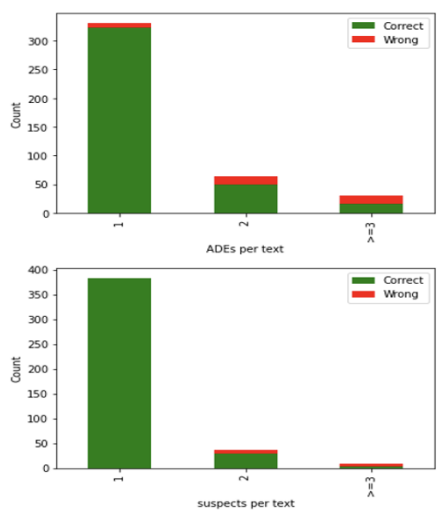

Figure 6: Performance of the model on ADE/Suspect extraction when the number of ADEs/Suspects per text varies.

## 7 Conclusion

In this paper, we propose a question answer based approach for solving ADE-suspect extraction problem by using a sequence-to-sequence transformer architecture, T5 (Raffel et al., 2020). We also detail our performance relative to the current baselines and present several experiments carried out utilizing various QA methodologies. We found that QA based RE approach outperforms existing baselines on the benchmark dataset. For industry usecases, it is recommended to use state-of-the-art NER followed by our QA based RE modeling for best results. This approach can be extended to extract ADEs and suspects from social media posts, clinical trial docs, medical transcripts, etc. We think that this work will be helpful when introducing a specific drug to the market or researching the negative effects of an existing drug since it will enable quick decisions to be made with little delay, preventing future causalities.

## 8 Future works

Although we used a sequence-to-sequence model in our study, Large Language Models (LLM) using decoder only transformers can also be used with the same methodology. With LLMs, we expect even better results than with the comparably small T5 models. An observed flaw in Approach 2 is that, it will still produce results even when there are no relationships between the drugs and ADEs in the text, which will impact the precision. To prevent this, we suggest to train the model to produce output like "no-suspect" and then post-process the predictions to remove them.